\documentclass[pdflatex,sn-nature]{sn-jnl}

\usepackage{graphicx}
\usepackage{amsmath,amssymb,amsfonts}
\usepackage{array}
\usepackage{booktabs}
\usepackage{longtable}
\usepackage{tabularx}
\usepackage{placeins}
\usepackage{xcolor}

\newcommand{\PMunit}{\ensuremath{\mu\mathrm{g}\,\mathrm{m}^{-3}}}

\hypersetup{bookmarksdepth=2,
  pdftitle={Interpretable PM2.5 Forecasting for Urban Air Quality: A Comparative Study of Adaptive Time-Series Models},
  pdfauthor={Moazzam Umer Gondal; Hamad Ul Qudous; Asma Ahmad Farhan; Sultan Alamri},
  pdfkeywords={PM2.5 forecasting; air quality; SARIMAX; Facebook Prophet; NeuralProphet; online residual correction}
}

\makeatletter
\let\PMabstractfont\abstractfont
\def\abstractfont{\PMabstractfont\rightskip=24pt plus 1fil}
\def\ps@submission{\let\@oddhead\@empty\let\@evenhead\@empty
\def\@oddfoot{\hfill\thepage\hfill}\let\@evenfoot\@oddfoot}
\makeatother

\input{glyphtounicode}\pdfgentounicode=1
\raggedright

\begin{document}
\pagestyle{submission}

\title[Interpretable PM\textsubscript{2.5} Forecasting]{Interpretable PM\textsubscript{2.5} Forecasting for Urban Air Quality: A Comparative Study of Adaptive Time-Series Models}

\author[1]{\fnm{Moazzam Umer} \sur{Gondal}}
\author[1]{\fnm{Hamad Ul} \sur{Qudous}}
\author*[1]{\fnm{Asma Ahmad} \sur{Farhan}}\email{asma.ahmad@nu.edu.pk}
\author[2]{\fnm{Sultan} \sur{Alamri}}

\affil[1]{\orgdiv{School of Computing}, \orgname{National University of Computer \& Emerging Sciences (FAST)}, \orgaddress{\city{Lahore}, \postcode{54000}, \country{Pakistan}}}
\affil[2]{\orgdiv{College of Computing and Informatics}, \orgname{Saudi Electronic University}, \orgaddress{\city{Riyadh}, \postcode{11673}, \country{Saudi Arabia}}}

\abstract{Balancing predictive performance, interpretation, and model updating is essential for accurate air-quality forecasting. We evaluate SARIMAX, Facebook Prophet, and NeuralProphet for hourly PM\textsubscript{2.5} in Beijing using API-derived estimates. Under Perfect Prognosis, weekly refitting and frozen parameters with EWMA residual correction are assessed using actual future gas concentrations. Validation and preprocessing using training data only precede a matched evaluation of 23 weekly origins and 3,624 observed hours. Weekly-refit pooled mean absolute errors are 30.43, 38.72, and 41.95~\PMunit{} for SARIMAX, Prophet, and NeuralProphet (seed 42), respectively. Predefined correction reduces frozen-model errors from 30.48 to 29.94, 43.93 to 37.69, and 53.39 to 37.62~\PMunit{}, respectively. Across three NeuralProphet seeds, frozen and refitted errors average 46.88 and 39.61~\PMunit{}, with standard deviations of 10.40 and 4.33, respectively. Correction benefits vary by seed, and the paired weekly interval for the SARIMAX improvement includes zero. Regressor effects and components give conditional interpretations, and different timing boundaries restrict speed comparisons. The results describe adaptation in a retrospective Perfect Prognosis evaluation; operational skill with forecast gas inputs and independently verified station targets remains untested.}


\maketitle
\thispagestyle{submission}

\section{Introduction}\label{sec1}

Air pollution is still a major environmental and public health problem worldwide. Air quality in large urban centers remains poor due to ongoing urbanization, industrial emissions, and motorized transport. Fine particulate matter (PM\textsubscript{2.5}) and inhalable particles (PM\textsubscript{10}) are of particular concern because they penetrate deeply into the respiratory system and contribute to respiratory and cardiovascular diseases \citep{bib1}. They also affect visibility and climate change.

This global challenge can be seen in cities such as Tehran and Beijing, where haze events are common and PM\textsubscript{2.5} concentrations are high, making real-time management challenging and highlighting the need for accurate forecasting \citep{bib2,bib3}.

Accurate predictions are crucial for public health and policymaking, as the WHO attributes more than 4 million premature deaths in 2019 to air pollution \citep{bib1}. Short-term prediction of PM\textsubscript{2.5} is essential for timely warnings and interventions, but it is challenging in megacities because of microclimates and complex interactions between pollutants \citep{bib4}. Smart city applications such as adaptive traffic and health management are also possible with robust predictive frameworks \citep{bib4,bib5}.

Research spans classical statistical methods, machine learning, deep learning, and hybrid decompositions. Recurrent networks, decomposition--reconstruction pipelines, and distributed learning have expanded the range of air-quality forecasting methods \citep{bib5,bib6,bib7,bib9,bib10,bib11}. Alongside predictive accuracy, repeated forecasting requires decisions about updating model parameters, incorporating recent observations, and managing computational cost. These decisions can be studied within established forecasting families without proposing a new architecture.

In this study, we examine hourly particulate matter forecasting for Beijing through a transparent comparison of SARIMAX, Facebook Prophet, and NeuralProphet. We examine how their performance changes under weekly expanding-window refitting and frozen parameters with an established EWMA bias correction. Experiments use Perfect Prognosis, which supplies actual future exogenous gases and isolates conditional forecasting and adaptation from the separate problem of predicting those gases. Our comparison is restricted to these three families, a predefined set of four gas inputs, and the evaluated period.

We make three contributions:
\begin{itemize}
\item An hourly forecasting workflow with chronological validation, explicit data-gap handling, and a common observed-hour evaluation mask under Perfect Prognosis.
\item A matched comparison of three time-series families over 23 weeks with weekly refitting and frozen parameters, supported by persistence references, predictor controls, lead-time errors, and repeated NeuralProphet seeds.
\item An evaluation of the benefits and constraints of online bias correction, with paired weekly uncertainty, fitted regressor effects, and separately defined computational measurements.
\end{itemize}

\section{Related work}\label{sec:related}

Classical time-series models offer explicit representations of temporal dependence and seasonality. ARIMA and neural-network comparisons in Beijing illustrate the distinction between linear and nonlinear formulations \citep{bib13}, while SARIMA--Prophet comparisons examine additive seasonal forecasting \citep{bib14}. Broader comparisons show that machine-learning methods can remain competitive with deep and additive models, depending on the region and forecast horizon \citep{bib15,bib16}. Tree ensembles, spatio-temporal feature engineering, and temporally weighted learning extend these approaches to richer inputs and changing conditions \citep{bib17,bib18,bib19,bib21}.

Deep and hybrid models address temporal, spatial, and multiscale relationships. Examples include attention-based Bi-LSTM estimation \citep{bib22}, optimized recurrent models \citep{bib23}, continuous-time neural formulations \citep{bib24}, and decomposition or architecture fusion \citep{bib7,bib27,bib28}. Recent BiGRU--1DCNN, imputation-based RNN--BiGRU, and CNN--BiLSTM--XGBoost studies further examine multi-station learning and missing-data treatment \citep{bibr1-1,bibr1-2,bibr1-3}. These methods answer different questions about representation and information use; their reported errors cannot establish a ranking under our protocol.

Recent Delhi studies provide more specific points of comparison. Sankar and Arasu combine wavelet extraction, PCA, hybrid feature optimization, and Bi-LSTM learning, using SHAP to describe feature contributions \citep{revWavelet}. Lakshmi and Krishnamoorthy use a bidirectional ConvLSTM encoder--decoder with spatial--temporal attention for multi-step PM$_{2.5}$ and PM$_{10}$ forecasting over horizons of 6--48 hours \citep{revSTA}. These studies emphasize feature representation and spatial information, whereas our single-location evaluation holds the primary input set fixed and examines repeated 168-hour forecasts under alternative update policies.

Temporal and cross-pollutant transfer also address adaptation. A winter-season TL-LSTM with multi-head attention combines temporal transfer with CorrXGBoost feature selection \citep{revWinter}; a subsequent centralized framework extends transfer to multiple pollutants and stations with feature-level and cross-pollutant attention \citep{revCrossPollutant}. Their adaptation and interpretability mechanisms differ from the fixed-parameter bias correction and explicit coefficients examined here. Distributed learning provides another route to sharing information across monitoring networks, with privacy, non-IID data, and communication constraints \citep{bib30,bib4}. Our experiments do not assess spatial transfer, federated training, or superiority over these architectures. They contribute a controlled comparison of updating strategies within three established families, with future-input availability and computational boundaries made explicit.

Table~\ref{tab:lr} summarizes these distinctions. Differences in datasets, target scaling, predictor availability, forecast horizons, and validation design prevent direct numerical comparison with the cited studies.

\begin{table}[htbp]
\caption{Selected literature and its relation to the present evaluation.}
\label{tab:lr}
\begingroup\small\setlength{\tabcolsep}{4pt}
\begin{tabularx}{\linewidth}{@{}l>{\raggedright\arraybackslash}X>{\raggedright\arraybackslash}X@{}}
\toprule
Study & Focus & Relation to this study \\
\midrule
\citep{bib14,bib15} & Statistical, additive, and ML comparisons & Motivates established model families; regions and protocols differ. \\
\citep{revWavelet} & Wavelet/PCA, optimized features, Bi-LSTM, SHAP & Feature optimization and explanation; our four-gas set is predefined. \\
\citep{revSTA} & Multi-station spatial--temporal attention; 6--48 h & Spatial representation and shorter horizons; our horizon is 168 h. \\
\citep{revWinter} & Winter temporal transfer and multi-head attention & Learned adaptation across seasons; our correction updates a scalar bias. \\
\citep{revCrossPollutant} & Centralized cross-pollutant transfer and attention & Multiple targets/stations; our target is one Beijing PM$_{2.5}$ series. \\
\citep{bib30} & Distributed/federated forecasting & Data locality and client heterogeneity are outside our experiment. \\
\botrule
\end{tabularx}\endgroup
\end{table}

\section{Methodology}\label{sec:methodology}

We formulate hourly PM\textsubscript{2.5} forecasting as a multistep prediction task with exogenous drivers. At forecast origin $o_w$, a model predicts $y_{o_w+h-1}$ for leads $h=1,\ldots,H$, where $H$ denotes the forecast horizon. Chronological partitioning and validation separate model development from subsequent evaluation. We compare parameter refitting with fixed-parameter forecasting and online residual correction; the specific experimental settings are given in Section~\ref{sec:implementation}.

Under Perfect Prognosis, future values of the exogenous regressors are treated as known over the forecast horizon. Future PM\textsubscript{2.5} is withheld from prediction and used only for scoring and subsequent updates. This assumption can approximate a setting with highly accurate external predictor forecasts, but those forecasts would introduce errors absent here. Without such externally available inputs, the evaluation remains an idealized conditional comparison and cannot establish operational forecast skill.

\subsection{Data preprocessing}\label{subsec:preprocessing}

Hourly timestamps are ordered on a regular calendar so that missing intervals retain their temporal meaning. Input validity and missingness are distinguished from high pollutant concentrations, which may represent genuine events. The original target scale is retained for interpretation and evaluation.

To avoid information leakage, transformations are estimated using data available before each fitting origin. Standardization expresses predictors relative to historical means and standard deviations. Missing-data handling must respect temporal order and distinguish fitting labels, inference history, and scoring targets; the model-specific policies are described in the implementation.

\subsection{\texorpdfstring{Predictor analysis}{Predictor analysis}}\label{subsec:feature_selection}

Pearson correlation and mutual information describe linear and potentially nonlinear associations between candidate predictors and PM\textsubscript{2.5}, using training data only. For continuous variables, their definitions are

\begin{equation}
\rho_{X,Y}=\frac{\mathrm{Cov}(X,Y)}{\sigma_X\sigma_Y}
\label{eq:pearson}
\end{equation}

\begin{equation}
I(X;Y)=\iint p(x,y)\log\!\left(\frac{p(x,y)}{p(x)p(y)}\right)\,dx\,dy
\label{eq:mi}
\end{equation}

Minimum redundancy--maximum relevance (mRMR) balances association with the target against dependence among selected predictors. In the relevance-to-redundancy quotient formulation (MIQ), candidate relevance $I(X_j;Y)$ is divided by its average mutual information with already selected predictors; the initial step uses relevance alone. These analyses describe associations, not causal effects. Strong target association indicates relevance and does not by itself establish redundancy among predictors.

\subsection{Forecasting models}\label{subsec:models}

We benchmark a classical state-space regression model (SARIMAX), a decomposable additive model (Facebook Prophet), and its autoregressive neural extension (NeuralProphet). Each supports multistep forecasting with external drivers under Perfect Prognosis.

\subsubsection{SARIMAX}

SARIMAX combines a regression on exogenous predictors with seasonal and nonseasonal temporal dependence. For predictor vector $x_t$, regression residual $u_t=y_t-\beta^{\mathsf T}x_t$, and lag operator $B$, a general multiplicative form is

\begin{equation}
\phi(B)\Phi(B^s)(1-B)^d(1-B^s)^D u_t=c+\theta(B)\Theta(B^s)\varepsilon_t
\label{eq:sarimax}
\end{equation}

Here $s$ is the seasonal period, $d$ and $D$ are nonseasonal and seasonal differencing orders, and $c$ is a constant where included. The polynomials $\phi$ and $\Phi$ represent autoregressive terms of orders $p$ and $P$, while $\theta$ and $\Theta$ represent moving-average terms of orders $q$ and $Q$. Their products retain seasonal/nonseasonal cross terms. The regression coefficients describe conditional associations given the temporal structure and other inputs; the selected specification is reported in Section~\ref{subsec:impl_models}.

\subsubsection{Facebook Prophet (FBP)}

Prophet represents a series through a trend, recurring seasonal patterns, and external-regressor contributions. In the additive formulation,

\begin{equation}
y_t=g(t)+s(t)+\beta^{\mathsf T}x_t+\varepsilon_t
\label{eq:prophet}
\end{equation}

The trend $g(t)$ can accommodate changepoints, while $s(t)$ represents periodic structure through Fourier terms. External regressors provide additional explanatory components. This decomposition permits inspection of trend, seasonal, and regressor contributions; the seasonalities and transformation choices used here are specified in the implementation.

\subsubsection{NeuralProphet (NP)}

NeuralProphet extends decomposable forecasting with learned autoregressive and regressor components. A representation combining trend, seasonality, lagged targets, and future-known inputs is

\begin{equation}
\hat y_{w,h}=T_{w,h}+S_{w,h}+A_h(y_{o_w-L:o_w-1})+F_h(x_{o_w:o_w+H-1})
\label{eq:neuralprophet}
\end{equation}

Here $L$ denotes the historical target-context length, $H$ the forecast horizon, and $h$ the forecast lead. The autoregressive contribution $A_h$ summarizes recent target history, while $F_h$ represents future-known regressors. The component parameterization and training configuration determine the flexibility of the model; their specific choices are reported in Section~\ref{subsec:impl_models}.

\subsection{Adaptive forecasting regimes}\label{subsec:regimes}

The two regimes differ in parameter updating while sharing forecast origins and evaluation criteria. Each may use observations revealed before the current origin.

\subsubsection{Regime 1: Weekly walk-forward refit}

At the beginning of week $w$, each model is fitted using the initial training period and the observed portions of earlier test weeks:

\begin{equation}
\mathcal D^{(w)}=\mathcal D_{\mathrm{train}}\cup\mathcal D_{1:w-1}
\label{eq:expanding}
\end{equation}

Model parameters and preprocessing are re-estimated from this expanding history. Earlier test outcomes may enter subsequent fitting, but the current forecast week's target remains unavailable. This regime incorporates newly revealed data at the cost of repeated parameter estimation.

\subsubsection{Regime 2: Frozen base model with online residual correction}

A base model is fitted once and retains its parameters and preprocessing during evaluation. Recent observations can still refresh its state or inference context. Adaptation is introduced through a bias applied to its forecast, with the model-specific information handling described in Section~\ref{subsec:impl_regimes}. The base multi-step residual for week $w$ and lead $h$ is

\begin{equation}
e_{w,h}=y_{o_w+h-1}-\hat y^{\mathrm{base}}_{w,h}
\label{eq:residual}
\end{equation}

Let $S_w$ contain the scored leads in week $w$. Apply the previously available bias $b_w$ before revealing that week's outcomes, then update it after the week using the observed base residual mean:

\begin{equation}
\hat y^{\mathrm{final}}_{w,h}=\hat y^{\mathrm{base}}_{w,h}+b_w,\qquad b_1=0
\label{eq:corrected}
\end{equation}

\begin{equation}
\bar e_w=\frac{1}{|S_w|}\sum_{h\in S_w}e_{w,h},\qquad b_{w+1}=\alpha\bar e_w+(1-\alpha)b_w
\label{eq:ewma}
\end{equation}

The smoothing factor $\alpha$ controls the weight assigned to the latest completed week's residual mean. Larger values respond more strongly to recent bias; $\alpha=1$ gives the previous-week mean-residual correction. EWMA is an established bias-tracking method applied within this comparison.

\subsection{Rolling evaluation protocol}\label{subsec:evaluation}

Forecasts are evaluated at consecutive weekly origins using a common observed-hour mask. Incomplete coverage is reported separately, and only original available target values enter scoring. This keeps the evaluated information and timestamps comparable across methods.

Pooled hourly MAE and RMSE give each scored hour equal weight:

\begin{equation}
\mathrm{MAE}=\frac{1}{N}\sum_{i=1}^{N}|y_i-\hat y_i|,\qquad \mathrm{RMSE}=\sqrt{\frac{1}{N}\sum_{i=1}^{N}(y_i-\hat y_i)^2}
\label{eq:metrics}
\end{equation}

Separately, mean weekly MAE and mean weekly RMSE give each origin equal weight; their sample standard deviations summarize variation across weeks. Mean weekly RMSE is not pooled RMSE. Lead-specific errors reveal how performance changes over the horizon. MAPE and SMAPE are omitted because near-zero concentrations make percentage errors unstable.

Persistence repeats the last observed target before an origin. Seasonal persistence repeats a preceding seasonal pattern across the forecast horizon. Such references use past targets only and therefore have a different information set from models supplied with future covariates under Perfect Prognosis.

\subsection{Statistical analysis}

Paired weekly error differences compare methods using the same origins and scoring masks. To address temporal dependence, moving-block bootstrap intervals retain contiguous groups of weekly differences. Negative A--B differences favor A. With a short dependent evaluation period, the intervals are descriptive and do not establish population-level significance. Variation across repeated training seeds is separated from temporal variation across forecast weeks.

\section{Implementation}\label{sec:implementation}

\begin{figure}[htbp]
\centering
\includegraphics[width=0.96\linewidth,keepaspectratio]{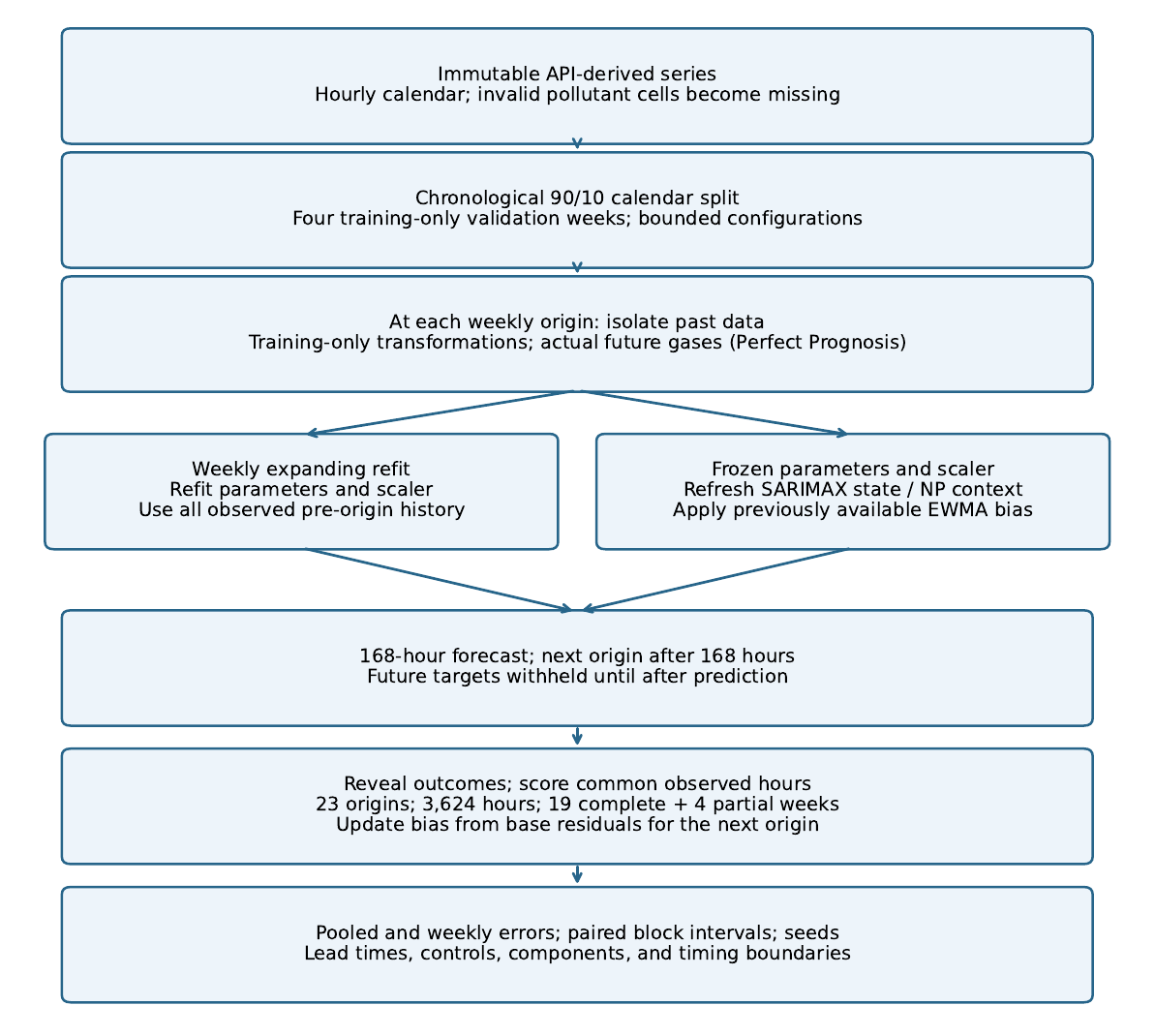}
\caption{Forecasting workflow. All fitting and preprocessing use history before each origin. Actual future covariates are supplied under Perfect Prognosis; future targets enter only scoring and subsequent updates.}
\label{fig:framework}
\end{figure}

All models were fitted on the same local AMD Ryzen 5 5500U system on CPU, with two configured model threads and approximately 14.96 GiB of reported system RAM; no GPU was used. The environment used Python 3.11.2, NumPy 1.26.4, pandas 2.3.1, scikit-learn 1.7.0, statsmodels 0.14.5, Prophet 1.1.7, NeuralProphet 0.9.0, and PyTorch 2.6.0. Descriptive mRMR used pymrmr 0.1.11 and figures used matplotlib. Model fitting, analysis, and notebook presentation are separate reproducible stages.

\subsection{Data assembly and preprocessing}\label{subsec:impl_data}

The concentrations of pollutants were gathered using the OpenWeather Air Pollution history API \citep{bibOpenWeather}, and temperature and dew point were gathered using the Open-Meteo ERA5 archive endpoint \citep{bibOpenMeteo}. These series are not observations from an identifiable monitoring station, but API-derived model or gridded estimates at a requested Beijing coordinate. Pollutants are reported in \PMunit{} and weather variables in degrees Celsius under provider defaults.

The source timestamps are sorted and reindexed on an hourly calendar without compressing gaps. Four chemically invalid predictor values of $-9999$ (two NO$_2$, one O$_3$, and one PM$_{10}$) are treated as missing in a working copy; the raw file and all PM$_{2.5}$ values are kept. Negative temperatures and dew points remain valid. No positive target extremes are removed or clipped during the primary evaluation. A separate sensitivity clips predictors only at the training set's 1st and 99th percentiles.

Input standardization uses means and standard deviations from complete predictor rows before the fitting origin. Frozen fits retain their initial scaler; refits recompute it from expanding history. Prophet fits observed complete target--predictor rows. SARIMAX retains missing targets in its state-space likelihood and forward-fills historical exogenous inputs using past values only. NeuralProphet trains on complete contiguous episodes of at least 336 hours, with shared model parameters and global target/time normalization. Its 168-lag/168-lead samples never cross a missing interval. Fitting and scoring targets are never imputed.

For inference requiring dense history, missing past values are filled causally from the value 168 hours earlier, recursively where necessary, with the latest preceding value as fallback. NeuralProphet consumes the latest 168-hour context; seasonal-persistence references also use this filled history. This policy changes computational context only. Missing future inputs receive temporary mean placeholders to construct a forecast, but unavailable-input hours are masked from saved predictions and scoring. Perturbing those placeholders to 10 standardized units leaves every scored prediction unchanged in the implemented models.

The requested coordinate was $(39.906217,116.3912757)$, stored in the CSV as $(39.9062,116.3913)$. The extraction uses the OpenWeather air-pollution history endpoint and Open-Meteo's ERA5 archive endpoint. The source contains 39,523 unique hourly rows from 25 November 2020 01:00 to 29 June 2025 19:00. Reindexing produces 40,267 hours with 744 absent timestamps across 21 intervals. The initial training period ends on 13 January 2025 00:00; testing begins one hour later. Training contains 36,240 calendar hours and 35,736 observed targets; testing contains 4,027 calendar hours and 3,787 observed targets.

Timestamps are interpreted as UTC based on the source's UNIX-second conversion and omitted weather time-zone option, but original execution logs and API responses are unavailable. Retrieval dates cannot be recovered reliably. The extraction used positional pollutant-component values rather than named lookup, so the historical field mapping cannot be independently authenticated. The CSV start also precedes the provider-documented historical start of 27 November 2020; this discrepancy remains unresolved. These limitations prevent treating the target as independently verified station truth.

The primary weekly evaluation ends on 23 June 2025 00:00. Nineteen weeks have 168 scored hours; the four partial weeks have 120, 120, 48, and 144. Table~\ref{tab:descriptive} describes original observed targets over the full training/test partitions, including the test remainder. All original observed targets are preserved; predictor-only clipping is evaluated separately from primary preprocessing.

\begin{table}[htbp]
\caption{Original observed PM$_{2.5}$ distributions. Concentrations are in \PMunit{}; SD denotes sample standard deviation and Q1--Q3 the interquartile interval.}
\label{tab:descriptive}
\begingroup\small\setlength{\tabcolsep}{4pt}
\begin{tabularx}{\linewidth}{@{}lrrrrrX@{}}
\toprule
Partition & $n$ & Mean & SD & Median & Q1--Q3 & Min--max \\
\midrule
Training & 35736 & 217.54 & 227.30 & 140.59 & 58.24--297.84 & 0.91--1825.93 \\
Test & 3787 & 155.34 & 216.57 & 72.02 & 27.68--163.12 & 0.00--1407.51 \\
\botrule
\end{tabularx}\endgroup
\end{table}

\subsection{\texorpdfstring{Predictor analysis}{Predictor analysis}}\label{subsec:impl_feature_selection}

Continuous MI is estimated with scikit-learn's nearest-neighbor estimator using seed 42. Descriptive mRMR uses the MIQ relevance-to-redundancy criterion after training-only quintile discretization, removing duplicate bin edges. At each selection step, candidate relevance $I(X_j;Y)$ is divided by its average mutual information with the already selected predictors; the initial step uses relevance alone. Five predictors are requested. These analyses describe associations, not causality.

The primary set of four gases $\{\mathrm{NO},\mathrm{NO}_2,\mathrm{CO},\mathrm{SO}_2\}$ is retained as a predefined subset to build on the original experiment. It is not claimed to be the optimal mRMR output. A broader nine-predictor set adds O$_3$, NH$_3$, temperature, dew point, and PM$_{10}$; no-input controls remove all exogenous variables. All frozen controls retain the selected model settings, so they assess input changes rather than separately tuned competitors.

Excluding PM$_{10}$ from the primary set keeps this comparison conditioned on gases rather than a contemporaneous particulate aggregate containing the fine fraction. PM$_{10}$ includes both fine and coarse particles, so it is not identical to PM$_{2.5}$ \citep{revPMFractions}. Its training correlation of 0.992 makes actual same-hour PM$_{10}$ a particularly close target proxy in this series. Leakage concerns depend on whether a feature is legitimate for the prediction task, not correlation alone \citep{revLeakage}. Actual future PM$_{10}$ would be unavailable to an operational forecast unless supplied externally, as would the actual future gases. The broad-set experiment is therefore an explicitly conditional Perfect Prognosis control, not evidence of operational skill. Lagged or independently forecast PM$_{10}$ could be legitimate inputs in a different protocol; the present exclusion is a scope restriction, not a claim that PM$_{10}$ is irrelevant.

\begin{figure}[htbp]
\centering
\includegraphics[width=0.96\linewidth,keepaspectratio]{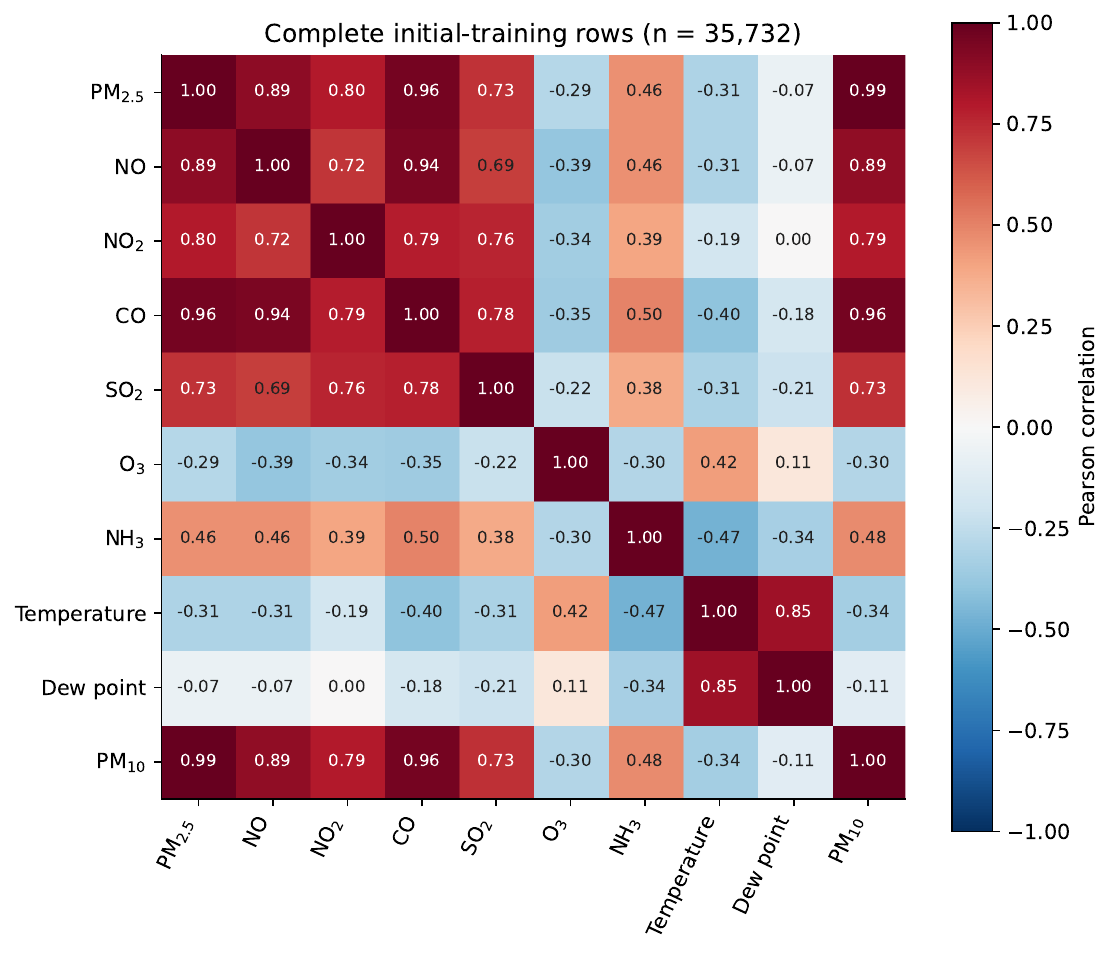}
\caption{Pearson correlations on 35,732 complete initial-training rows after invalid-input masking. Associations do not establish causal effects.}
\label{fig:corr_heatmap}
\end{figure}

\begin{figure}[htbp]
\centering
\includegraphics[width=0.96\linewidth,keepaspectratio]{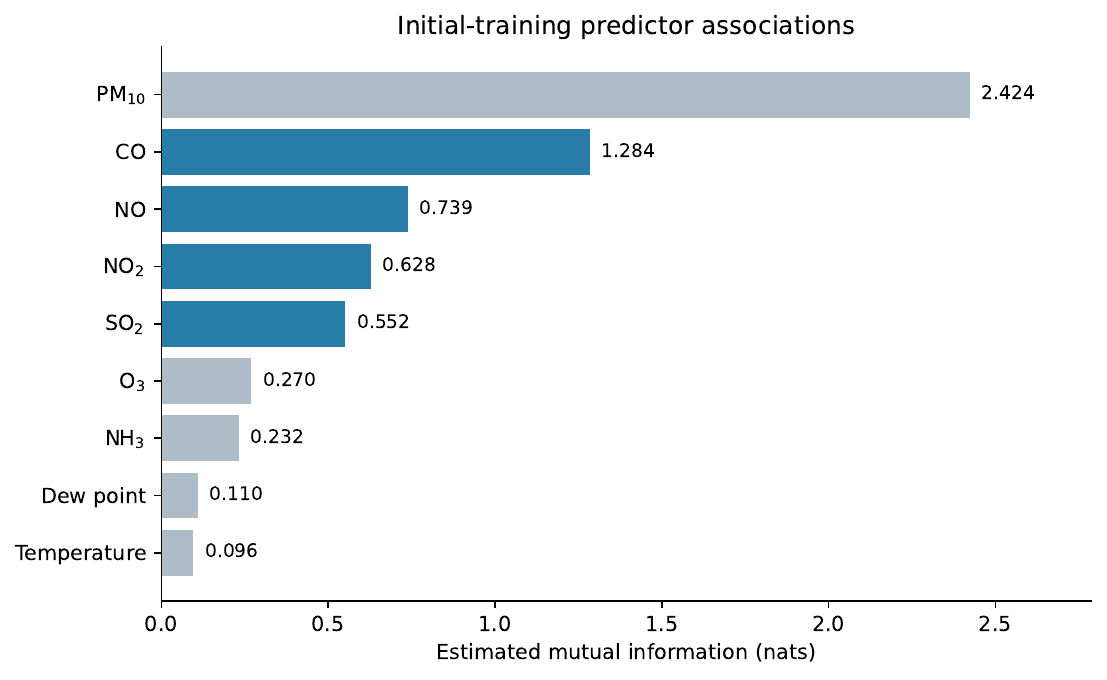}
\caption{Training-only continuous mutual information between each candidate predictor and the original target. These descriptive scores do not select the primary subset of four gases.}
\label{fig:mi_bar}
\end{figure}

Training-only complete-row analysis uses 35,732 hours. PM$_{10}$ has the largest Pearson correlation and MI with PM$_{2.5}$, followed by CO and NO (Table~\ref{tab:features}; Figs.~\ref{fig:corr_heatmap}--\ref{fig:mi_bar}). MIQ mRMR ranks PM$_{10}$, dew point, NO$_2$, NO, and CO as its first five predictors. This differs from the primary subset of four gases and does not validate that subset as optimal. High target association is relevance, not evidence of redundancy among predictors; the broader-input control examines the consequences of excluding PM$_{10}$ and other candidates.

\begin{table}[htbp]
\caption{Training-only predictor rankings. MI is estimated on continuous inputs; mRMR uses separately discretized inputs.}
\label{tab:features}
\begingroup\small\setlength{\tabcolsep}{4pt}
\begin{tabularx}{\linewidth}{@{}Xrrrr@{}}
\toprule
Predictor & Pearson $r$ & MI & Absolute-$r$ rank & MI rank \\
\midrule
NO & 0.892 & 0.739 & 3 & 3 \\
NO$_2$ & 0.796 & 0.628 & 4 & 4 \\
CO & 0.957 & 1.284 & 2 & 2 \\
SO$_2$ & 0.726 & 0.552 & 5 & 5 \\
O$_3$ & -0.287 & 0.270 & 8 & 6 \\
NH$_3$ & 0.457 & 0.232 & 6 & 7 \\
Temperature & -0.309 & 0.096 & 7 & 9 \\
Dew point & -0.067 & 0.110 & 9 & 8 \\
PM$_{10}$ & 0.992 & 2.424 & 1 & 1 \\
\botrule
\end{tabularx}\endgroup
\end{table}

\subsection{Model configuration}\label{subsec:impl_models}

For SARIMAX, the selected specification sets $p=q=P=Q=1$, $d=D=0$, $s=24$, and $c=0$ in Eq.~\ref{eq:sarimax}; the multiplicative seasonal/nonseasonal cross terms are retained.

Daily, weekly, and yearly seasonalities and default changepoint handling are retained. Additional regressor standardization is disabled because inputs have already been transformed using training-only statistics.

The configuration uses 168 lags and 168 direct forecast leads, daily/weekly/yearly seasonalities, and default linear autoregressive and future-regressor modules. Events, holidays, lagged exogenous regressors, and hidden autoregressive layers are not added. Global normalization is learned from the retained pre-origin training episodes and predictions are returned to the original concentration scale. Finite training losses and exact origin/lead extraction are checked; completing the specified epochs does not prove optimization to a global minimum.

Time-ordered validation uses four consecutive weekly origins from 16 December 2024 01:00 through 6 January 2025 01:00, all before testing. The screening base fit precedes the first validation origin and produces the subsequent weekly forecasts with frozen parameters and refreshed state/context. The bounded grid contains four SARIMAX combinations with $p=q=P=Q=1$, $d,D\in\{0,1\}$, and period 24; additive/multiplicative Prophet seasonality; and 30/50 NeuralProphet epochs. Candidates are selected by mean weekly MAE, with pooled RMSE as tie-breaker. Seven of eight candidates complete; SARIMAX $(1,0,1)\times(1,1,1,24)$ fails and is excluded. This limited winter validation is not an exhaustive hyperparameter search.

\begin{table}[htbp]
\caption{Selected training-only configurations, used unchanged in both regimes.}
\label{tab:configuration}
\begingroup\small\setlength{\tabcolsep}{4pt}
\begin{tabularx}{\linewidth}{@{}l>{\raggedright\arraybackslash}Xr@{}}
\toprule
Family & Configuration & Validation MAE \\
\midrule
SARIMAX & $(1,0,1)	imes(1,0,1,24)$; no constant; stationarity/invertibility constraints disabled & 14.99 \\
Prophet & Additive; daily/weekly/yearly seasonalities; default trend/changepoints; no additional regressor scaling & 28.52 \\
NeuralProphet & 168 lags/leads; 50 epochs; batch 128; learning rate 0.001; no early stopping; global normalization & 27.59 \\
\botrule
\end{tabularx}\endgroup
\end{table}

NeuralProphet uses seeds 42, 123, and 2026 for both regimes; validation and predictor controls use seed 42. Training losses, episode sample counts, forecast timestamps, and the 168 extracted leads are checked. SARIMAX's test fits use an earlier converged, corrected training fit for initialization and complex-step L-BFGS-B with at most 200 iterations and 100 line searches. Acceptance requires finite parameters, solver convergence, and nondecreasing training likelihood, without consulting test errors. The broad-input SARIMAX control also receives one bounded 400-iteration conditional-sum-of-squares initialization attempt, which remains nonconverged; no predictions from this failed control are scored.

\subsection{Adaptive regime implementation}\label{subsec:impl_regimes}

In the frozen regime, SARIMAX filters newly revealed history without refitting, NeuralProphet receives the latest 168-hour target context, and Prophet evaluates the actual future dates and covariates. Walk-forward refits re-estimate their parameters and scaler using all observed pre-origin history. The first-origin fit and forecast are identical to the corresponding frozen fit and are reused only after exact configuration and signature checks.

The primary smoothing factor $\alpha=0.3$ is predefined. Sensitivities use 0.1, 0.2, 0.5, 0.7, and 1.0; the latter is the previous-week mean-residual correction. No value is selected from test performance. Four partial weeks update the primary bias using their observed scoring hours. EWMA is an established bias-tracking method applied here, rather than a new algorithm; no separate Kalman bias-correction alternative is evaluated.

Walk-forward refits use all available pre-origin history at each week. Frozen streams retain their initial scaler and parameters while refreshing the model-specific state/context described above. EWMA is applied before outcomes and updated afterwards from base residuals. The common observed-only score mask is identical across models, regimes, seeds, and controls. The strict 16-week sensitivity carries the bias forward across excluded weeks and updates only after eligible weeks, whereas the primary 23-week policy also updates after partially observed weeks.

\begin{table}[htbp]
\caption{Weekly forecasting sequence and information boundaries.}
\label{tab:pseudocode}
\begingroup\small\setlength{\tabcolsep}{4pt}
\begin{tabularx}{\linewidth}{@{}l>{\raggedright\arraybackslash}X@{}}
\toprule
Step & Operation \\
\midrule
1 & At origin $o_w$, isolate history with timestamps earlier than $o_w$ and obtain future covariates under Perfect Prognosis. \\
2 & Refit parameters/scaler from history (Regime 1), or retain them and refresh state/context (Regime 2). \\
3 & Produce 168 aligned base predictions; apply $b_w$ for frozen correction before current outcomes are revealed. \\
4 & After the week, score only shared observed target/covariate hours; retain partial-week coverage counts. \\
5 & Use base residuals from scored hours to update $b_{w+1}$; make revealed history available at the next origin. \\
\botrule
\end{tabularx}\endgroup
\end{table}

\subsection{\texorpdfstring{Evaluation and statistical settings}{Evaluation and statistical settings}}\label{subsec:impl_evaluation}

The complete hourly calendar is split chronologically into 90\% initial training and 10\% testing. Forecast horizon and origin step are both 168 hours.

The primary schedule retains all 23 complete calendar weeks (3,864 scheduled hours), including four partially observed weeks. A shared mask requires the original target and all broad-set covariates to be observed, producing 3,624 scored hours for every method. The final 163-hour calendar remainder is reported but excluded from the fixed 168-hour evaluation. A secondary sensitivity uses 16 weeks with fully observed future and prior 168-hour context (2,688 hours); it is not the primary result.

Paired weekly MAE differences use identical origins and scoring masks. Descriptive 95\% intervals are obtained from 2,000 moving-block bootstrap replicates with a primary block length of three weeks and sensitivities of two and four weeks. NeuralProphet paired losses average the three seed-level weekly losses, rather than averaging forecasts. Negative A--B differences favor A. Intervals are not adjusted for multiple comparisons and rely on only 23 dependent weeks; they do not establish broad population-level significance. Seed means and sample standard deviations are reported separately from temporal dispersion.

The test period had already been examined during model development. The present evaluation is therefore retrospective and cannot be treated as an untouched confirmatory test. The input-validity rule, candidate grid, primary scoring coverage, and EWMA factor were specified before inspecting the reported test results.

\subsection*{Code availability}\label{sec:code_availability}

The code supporting this study is archived on Zenodo (version 1.0.0): \url{https://doi.org/10.5281/zenodo.23170178}.

\section{Results}\label{sec:results}

All primary results use 23 origins and 3,624 scored hours. Pooled hourly errors and equally weighted weekly mean errors are reported separately; dispersion denotes sample SD across the 23 weeks. NeuralProphet uses seed 42 in the main family comparison, with all seeds reported subsequently. Every primary SARIMAX refit completed all 23 origins. All errors are in \PMunit{}.

\subsection{Regime 1: Weekly walk-forward refitting}\label{subsec:results_regime1}

\begin{figure}[htbp]
\centering
\includegraphics[width=0.96\linewidth,keepaspectratio]{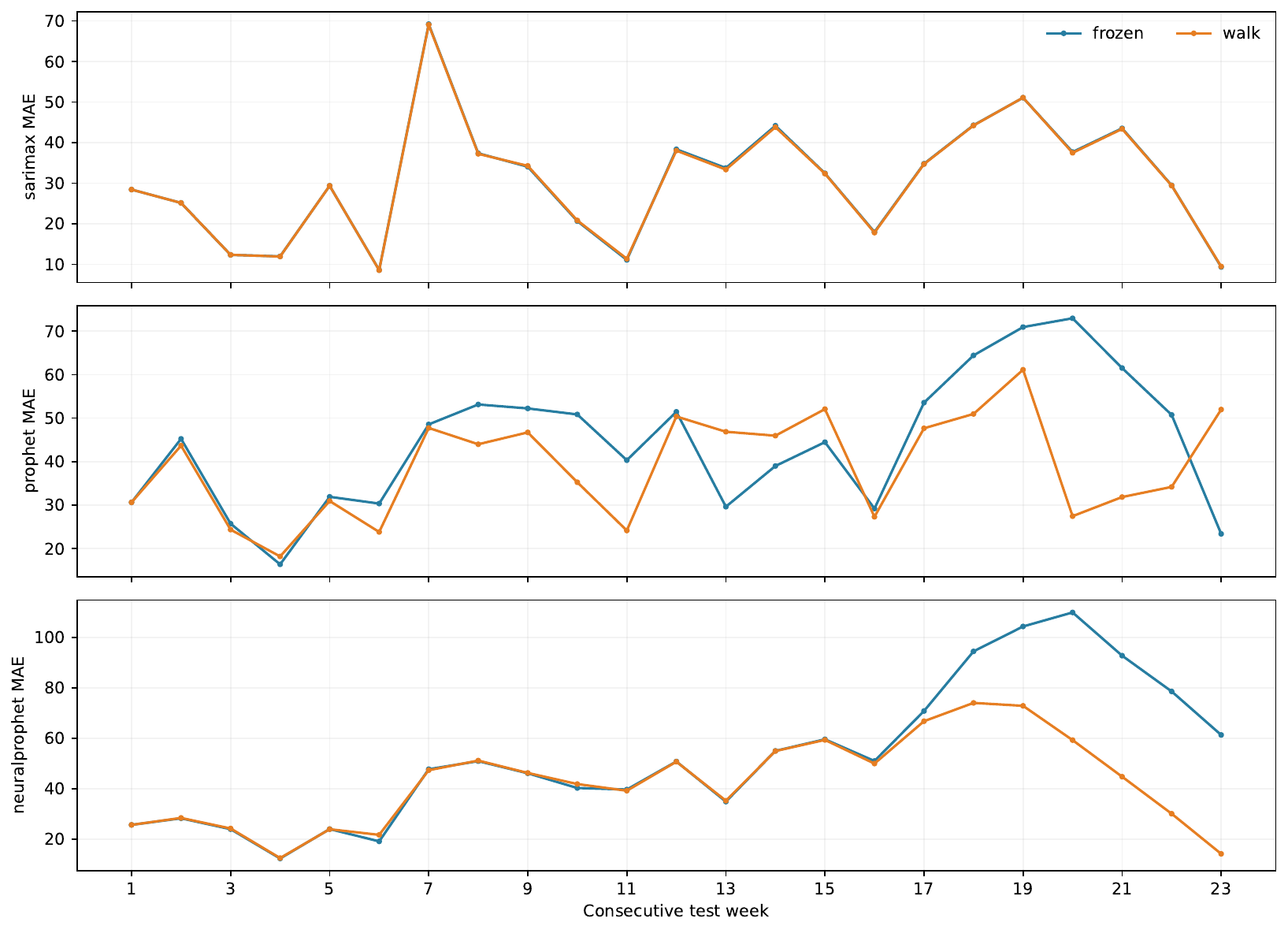}
\caption{Chronological weekly MAE comparison of frozen base forecasts and weekly refitting on 23 origins and 3,624 shared scoring hours. NeuralProphet uses seed 42.}
\label{fig:regimes}
\end{figure}

\begin{figure}[htbp]
\centering
\includegraphics[width=0.96\linewidth,keepaspectratio]{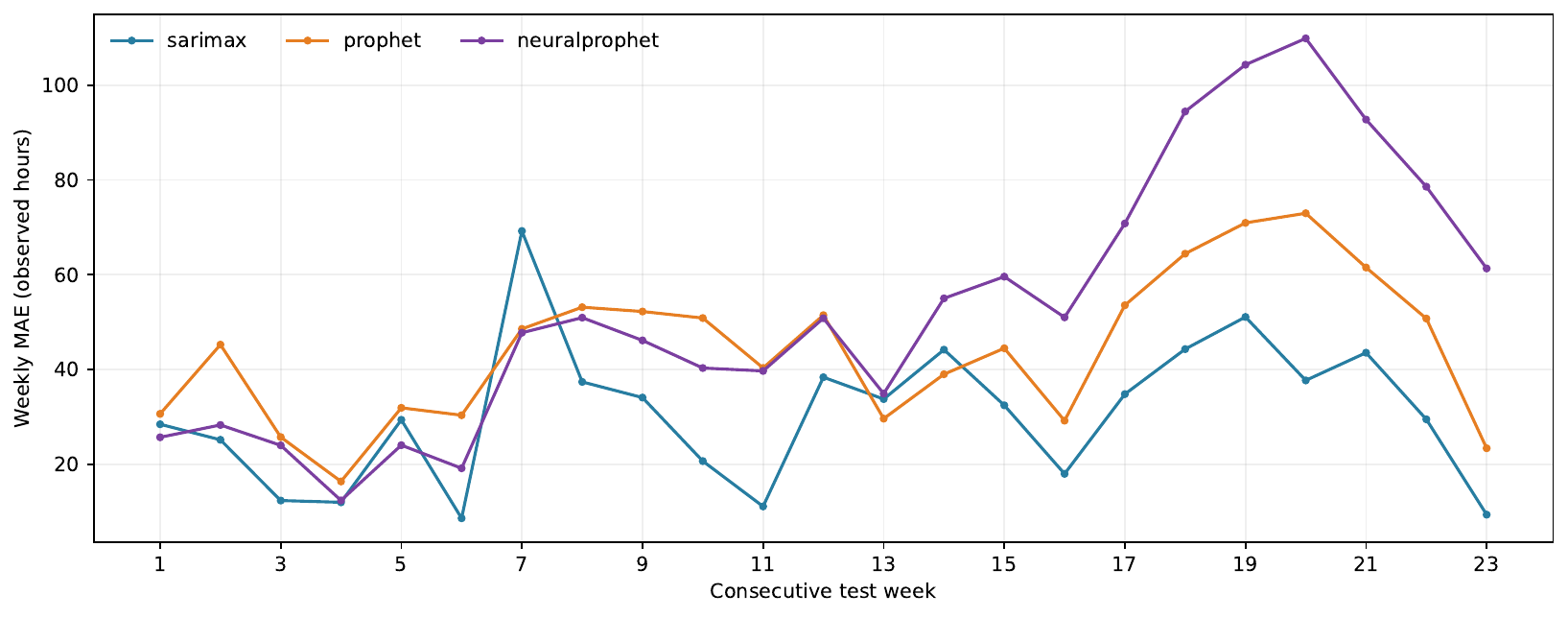}
\caption{Chronological weekly MAE of frozen base forecasts for all 23 origins. NeuralProphet uses seed 42; partial weeks contribute only their observed scoring hours.}
\label{fig:chronology}
\end{figure}

SARIMAX has the lowest pooled MAE and RMSE in the matched walk-forward comparison (30.43 and 46.88), followed by Prophet (38.72 and 51.13) and seed-42 NeuralProphet (41.95 and 56.51). Weekly MAE dispersion remains substantial for every family (Table~\ref{tab:regime1_results}). Figure~\ref{fig:regimes} compares the two parameter-update regimes, and Fig.~\ref{fig:chronology} shows every evaluation week rather than selected extremes.

\begin{table}[htbp]
\caption{Weekly walk-forward results on the common observed-hour mask. Weekly and pooled aggregations are distinct.}
\label{tab:regime1_results}
\begingroup\small\setlength{\tabcolsep}{4pt}
\begin{tabularx}{\linewidth}{@{}Xrrrr@{}}
\toprule
Model & Pooled MAE & Pooled RMSE & Weekly MAE $\pm$ SD & Weekly RMSE $\pm$ SD \\
\midrule
SARIMAX & 30.43 & 46.88 & 30.60 $\pm$ 14.92 & 41.66 $\pm$ 21.78 \\
Prophet & 38.72 & 51.13 & 39.01 $\pm$ 11.80 & 49.15 $\pm$ 14.89 \\
NP (42) & 41.95 & 56.51 & 42.38 $\pm$ 17.80 & 52.54 $\pm$ 21.79 \\
\botrule
\end{tabularx}\endgroup
\end{table}

\subsection{Regime 2: Frozen forecasting with online residual correction}\label{subsec:results_regime2}

\begin{figure}[htbp]
\centering
\includegraphics[width=0.96\linewidth,keepaspectratio]{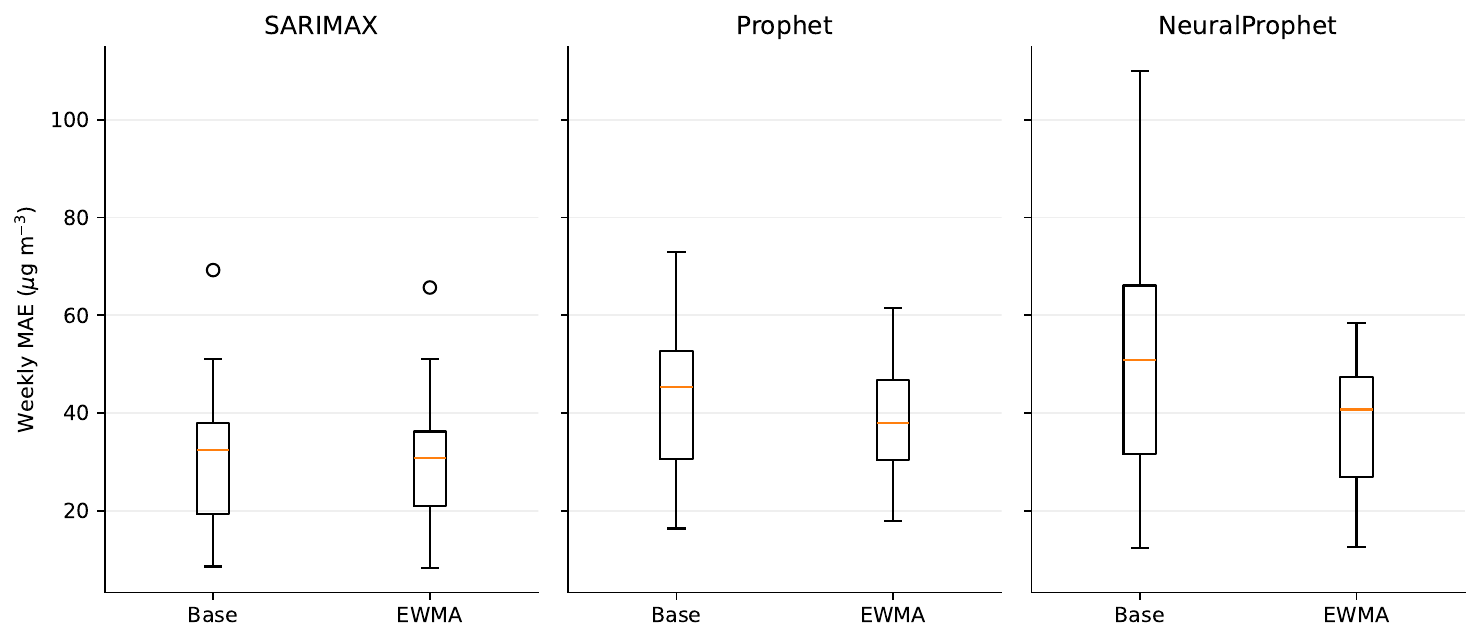}
\caption{Weekly MAE distributions for frozen base and EWMA-corrected models over 23 origins. NeuralProphet uses seed 42; the plot summarizes temporal variation rather than uncertainty across training seeds.}
\label{fig:weekly_distribution}
\end{figure}

With $\alpha=0.3$, frozen SARIMAX has the lowest pooled corrected MAE (29.94) and RMSE (43.80). Prophet's MAE decreases from 43.93 to 37.69, while seed-42 NeuralProphet decreases from 53.39 to 37.62 (Table~\ref{tab:regime2_results}). These are differences in pooled hourly MAE; paired weekly differences below use equal origin weights. Figure~\ref{fig:weekly_distribution} shows the distributions of frozen and corrected weekly errors.

\begin{table}[htbp]
\caption{Frozen base and EWMA-corrected results. Correction uses the previously available bias, with predefined $\alpha=0.3$.}
\label{tab:regime2_results}
\begingroup\small\setlength{\tabcolsep}{4pt}
\begin{tabularx}{\linewidth}{@{}Xrrrr@{}}
\toprule
Model & Pooled MAE & Pooled RMSE & Weekly MAE $\pm$ SD & Weekly RMSE $\pm$ SD \\
\midrule
SAR base & 30.48 & 46.96 & 30.66 $\pm$ 14.99 & 41.74 $\pm$ 21.85 \\
SAR + EWMA & 29.94 & 43.80 & 30.17 $\pm$ 13.13 & 39.60 $\pm$ 19.07 \\
FBP base & 43.93 & 57.56 & 44.20 $\pm$ 15.38 & 54.70 $\pm$ 18.51 \\
FBP + EWMA & 37.69 & 49.97 & 38.07 $\pm$ 10.84 & 48.27 $\pm$ 14.08 \\
NP (42) base & 53.39 & 70.17 & 53.12 $\pm$ 27.68 & 63.23 $\pm$ 29.65 \\
NP + EWMA & 37.62 & 50.91 & 38.29 $\pm$ 12.57 & 48.58 $\pm$ 17.20 \\
\botrule
\end{tabularx}\endgroup
\end{table}

\subsection{\texorpdfstring{Paired comparisons and seed variability}{Paired comparisons and seed variability}}

\begin{table}[htbp]
\caption{Paired mean weekly MAE differences and descriptive 95\% moving-block intervals (2,000 replicates; block length three). Negative values favor the first method; NP losses average three seeds.}
\label{tab:paired}
\begingroup\small\setlength{\tabcolsep}{4pt}
\begin{tabularx}{\linewidth}{@{}Xrr@{}}
\toprule
Comparison & Difference & 95\% interval \\
\midrule
SAR EWMA -- base & -0.49 & [-3.91, 2.25] \\
FBP EWMA -- base & -6.13 & [-14.01, -0.99] \\
NP EWMA -- base & -9.25 & [-18.98, -0.25] \\
SAR EWMA -- FBP EWMA & -7.90 & [-11.33, -3.63] \\
FBP EWMA -- FBP refit & -0.94 & [-1.90, 0.73] \\
\botrule
\end{tabularx}\endgroup
\end{table}

The interval for SARIMAX's small EWMA improvement includes zero. Prophet's correction interval excludes zero for the primary and two-/four-week block lengths. The seed-averaged NeuralProphet correction interval excludes zero with a three-week block but includes zero with a four-week block. Corrected Prophet and refitted Prophet have a paired weekly difference of $-0.94$ with an interval spanning zero; this is not an equivalence test.

\begin{table}[htbp]
\caption{NeuralProphet pooled MAE by training seed. These are individual runs, not an ensemble.}
\label{tab:seeds}
\begingroup\small\setlength{\tabcolsep}{4pt}
\begin{tabularx}{\linewidth}{@{}Xrrr@{}}
\toprule
Seed & Frozen base & Weekly refit & Frozen + EWMA \\
\midrule
42 & 53.39 & 41.95 & 37.62 \\
123 & 52.37 & 42.25 & 38.02 \\
2026 & 34.89 & 34.61 & 35.46 \\
\botrule
\end{tabularx}\endgroup
\end{table}

Across seeds, mean pooled MAE is $46.88\pm10.40$ for frozen forecasts and $39.61\pm4.33$ for refits (mean $\pm$ sample SD). Seed 2026 is more accurate than seeds 42/123 before correction; its corrected MAE slightly increases from 34.89 to 35.46. Thus, these results do not establish an inherent NeuralProphet weakness or a uniform correction benefit. Normalization, contiguous training episodes, and lead extraction were audited; remaining seed variation is observed, while explanations involving optimization or seasonal generalization remain tentative.

\subsection{\texorpdfstring{Lead time, concentration, and sensitivity}{Lead time, concentration, and sensitivity}}\label{subsec:lead}

\begin{figure}[htbp]
\centering
\includegraphics[width=0.96\linewidth,keepaspectratio]{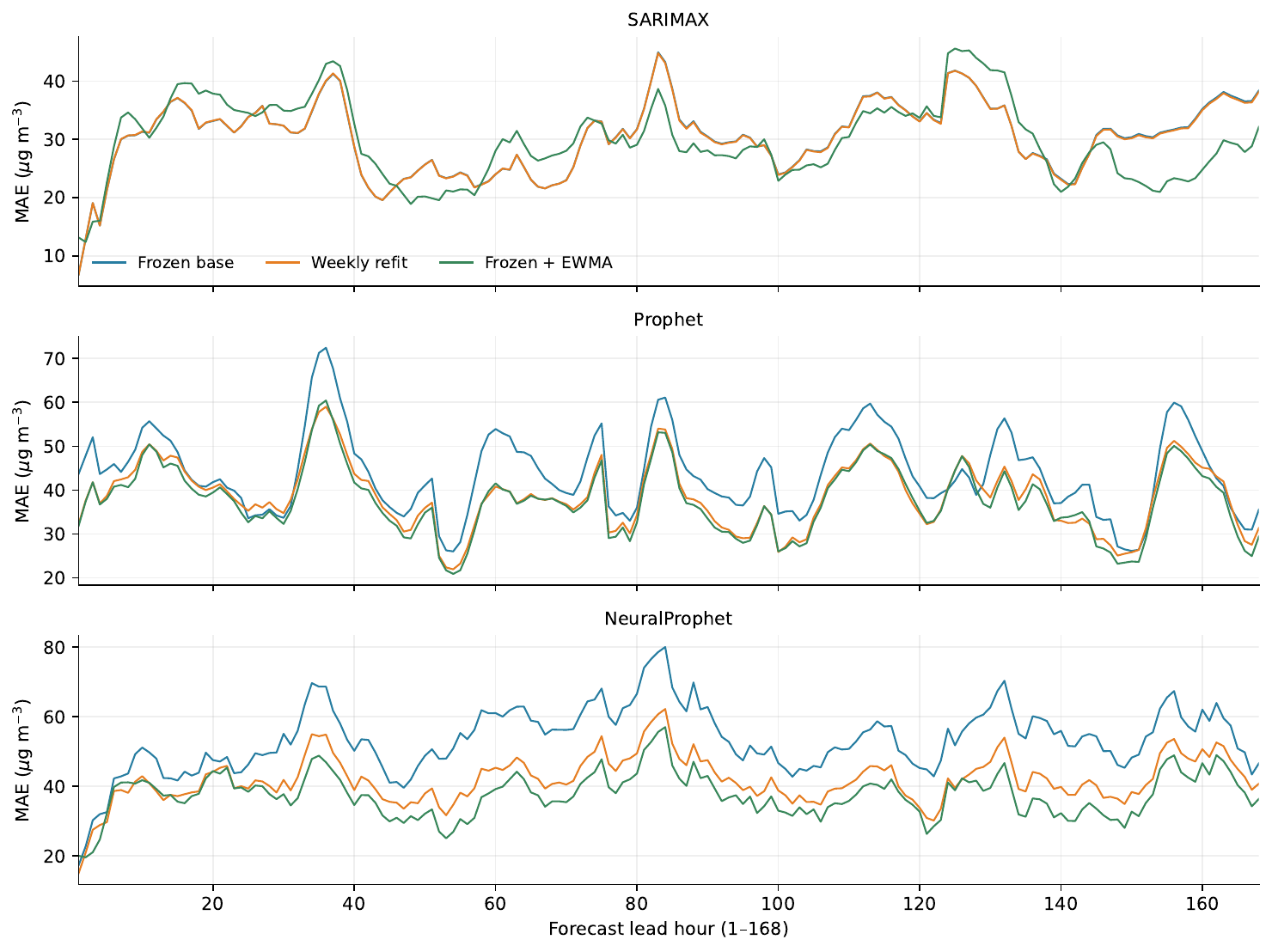}
\caption{MAE by hourly lead for frozen base forecasts, weekly refits, and frozen EWMA correction. NeuralProphet uses seed 42; lead-specific scoring counts reflect observed-hour coverage.}
\label{fig:lead_hour}
\end{figure}

\begin{figure}[htbp]
\centering
\includegraphics[width=0.96\linewidth,keepaspectratio]{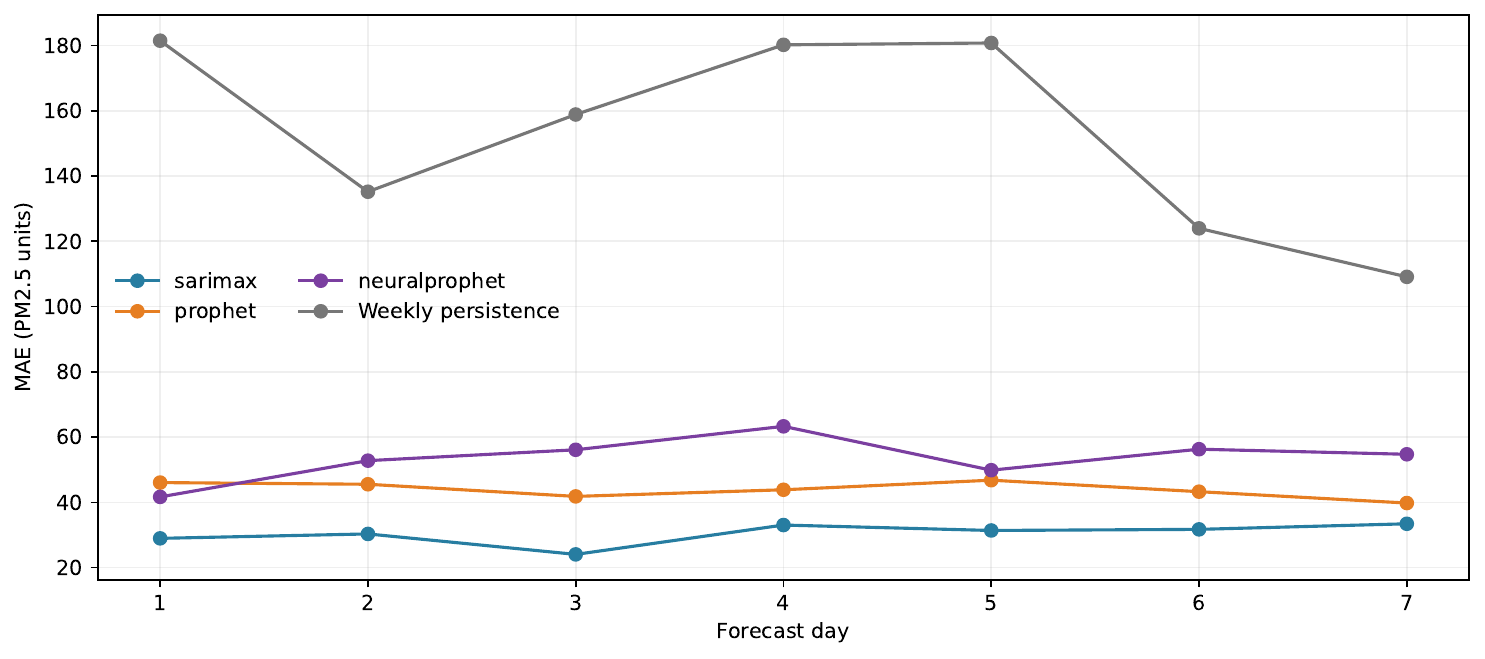}
\caption{MAE by forecast day over the seven-day horizon, using matched observed hours. NeuralProphet uses seed 42.}
\label{fig:lead_day}
\end{figure}

Figures~\ref{fig:lead_hour} and \ref{fig:lead_day} show variation over the 168-hour horizon. For each lead, errors are averaged over the matched observed targets, whose number may vary because of missing hours. The high-concentration threshold is the training 95th percentile, 691.34~\PMunit{}. Frozen seed-42 NeuralProphet has MAE 32.32 on 158 scored hours in six weeks, while SARIMAX and Prophet have MAEs of 51.02 and 50.60, respectively. This reverses the aggregate base-model ordering and prevents claims of uniformly better performance.

\begin{figure}[htbp]
\centering
\includegraphics[width=0.96\linewidth,keepaspectratio]{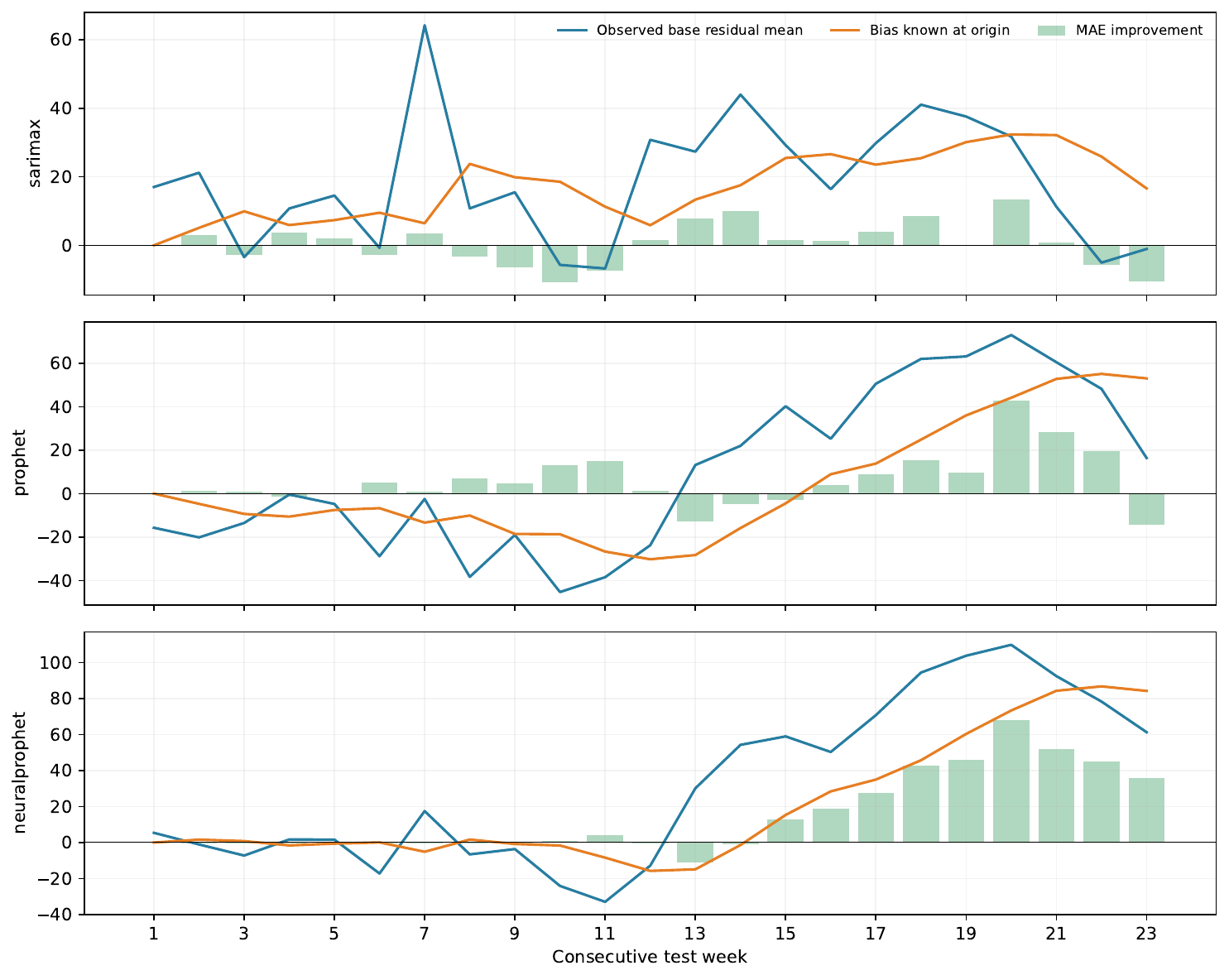}
\caption{Frozen-model correction diagnostics for the predefined $\alpha=0.3$. Residuals are base-model residuals; bias is available before the forecast week and updated only afterwards. NeuralProphet seed-specific behavior is reported in the text and tables.}
\label{fig:correction}
\end{figure}

The predefined alpha sensitivity shows that correction response varies between families and seeds. The base residuals, available bias, and correction benefit for seed 42 are displayed in Figure~\ref{fig:correction}. For SARIMAX, equally weighted weekly MAE is 29.50 at $\alpha=0.1$, 30.17 at 0.3, and 31.33 at 1.0, versus 30.66 without correction. Prophet improves over its base at all evaluated nonzero factors, while seed-2026 NeuralProphet is slightly worse at those factors. These descriptive test sensitivities are not used to select an alternative alpha. A bias calculated from the previous week can help a persistent offset but may overcorrect when the level changes.

The base forecasts for the secondary 16-week complete-context comparison are unchanged. The strict-policy corrected MAEs are 29.09 for SARIMAX, 39.16 for Prophet, and 38.67 for seed-42 NeuralProphet. On those same hours, the primary partial-week update policy gives 29.15, 35.78, and 34.26, respectively. The differences indicate that both week inclusion and bias-update opportunities are important; the 23-week results remain primary.

\subsection{\texorpdfstring{Persistence references and predictor controls}{Persistence references and predictor controls}}

\begin{table}[htbp]
\caption{Frozen predictor controls and persistence references: pooled errors on the same 3,624 scored hours. NP controls use seed 42; broad-input SARIMAX is unavailable because of nonconvergence.}
\label{tab:controls}
\begingroup\small\setlength{\tabcolsep}{4pt}
\begin{tabularx}{\linewidth}{@{}Xrr@{}}
\toprule
Method & MAE & RMSE \\
\midrule
Last observed value & 125.74 & 198.33 \\
Daily persistence & 121.52 & 202.34 \\
Weekly persistence & 152.90 & 248.65 \\
SAR no inputs & 115.24 & 173.53 \\
FBP no inputs & 134.07 & 190.40 \\
NP no inputs & 119.50 & 177.02 \\
FBP broad inputs & 15.87 & 27.57 \\
NP broad inputs & 22.35 & 33.29 \\
SAR clipped inputs & 33.17 & 51.55 \\
FBP clipped inputs & 47.68 & 62.76 \\
NP clipped inputs & 58.28 & 75.95 \\
\botrule
\end{tabularx}\endgroup
\end{table}

The four-gas models outperform the target-only references on aggregate, but receive actual future gases unavailable to those references. No-input controls retain the selected settings and have MAEs of 115.24, 134.07, and 119.50 for SARIMAX, Prophet, and NeuralProphet. Broad-input Prophet and NeuralProphet improve to 15.87 and 22.35 with actual future PM$_{10}$ among their inputs. This result highlights the information gained from additional predictors and argues against claiming the four gases are optimal. Because five predictors are added together, the control does not isolate PM$_{10}$'s individual contribution. Its contemporaneous particulate input also changes the conditioning information, as explained in Section~\ref{subsec:impl_feature_selection}. Broad-input SARIMAX fails both bounded optimizer attempts, preventing a complete broad-input family ranking. Predictor-only clipping increases MAE for all three selected-input frozen models; high target concentrations remain unmodified.

\subsection{\texorpdfstring{Model interpretation and computational measurements}{Model interpretation and computational measurements}}\label{subsec:interpretation}

\begin{table}[htbp]
\caption{Initial frozen-model regressor coefficients, expressed as a PM$_{2.5}$ change in \PMunit{} per one training-input standard deviation, conditional on other inputs and temporal terms.}
\label{tab:effects}
\begingroup\small\setlength{\tabcolsep}{4pt}
\begin{tabularx}{\linewidth}{@{}Xrr@{}}
\toprule
Input & SARIMAX & Prophet \\
\midrule
NO & -36.62 & -46.32 \\
NO$_2$ & 5.17 & 35.40 \\
CO & 237.06 & 265.48 \\
SO$_2$ & 3.22 & -24.94 \\
\botrule
\end{tabularx}\endgroup
\end{table}

The frozen coefficients are shown in Table~\ref{tab:effects}. Because refit input scales change, stability is assessed using coefficients converted to raw-input units. Across 23 refits, SARIMAX's NO coefficient ranges from $-0.1996$ to $-0.1992$ and CO from 0.09596 to 0.09598; Prophet's corresponding ranges are $-0.2519$ to $-0.2410$ and 0.1061 to 0.1075. Positive marginal correlations alongside negative conditional NO coefficients illustrate the effect of correlated regressors and model conditioning, rather than a causal protective effect.

For frozen Prophet, component means weight each of the 3,624 scored hours equally. Mean absolute contributions are 237.53 for CO, 43.55 for NO$_2$, 38.45 for NO, 25.02 for SO$_2$, 221.48 for trend, and 19.18, 4.05, and 10.12 for daily, weekly, and yearly seasonality. Signed contributions may cancel, so component magnitude is not an independent feature-importance measure. These outputs demonstrate explicit model interpretation without identifying emission mechanisms.

\begin{table}[htbp]
\caption{New walk-forward fit times for origins 2--23, excluding preprocessing and prediction. Boundaries differ across implementations and do not support a controlled speed ratio.}
\label{tab:timing}
\begingroup\small\setlength{\tabcolsep}{4pt}
\begin{tabularx}{\linewidth}{@{}lrlrX@{}}
\toprule
Family & Fits & Timer boundary & Median (s) & Range (s) \\
\midrule
SARIMAX & 22 & Optimization only & 110.13 & 51.18--215.06 \\
Prophet & 22 & Including initialization/import & 23.91 & 16.22--33.74 \\
NeuralProphet & 66 & Including initialization/import & 182.02 & 170.63--202.59 \\
\botrule
\end{tabularx}\endgroup
\end{table}

Fit timings in Table~\ref{tab:timing} are separated by boundary. Preprocessing, SARIMAX state updates, model prediction, and verification are recorded separately; first-origin reused forecasts retain their source computation cost. Frozen correction arithmetic takes approximately 0.002 seconds for a complete stream in repeated isolated measurements, excluding fitting, CSV I/O, and plotting. Recorded execution attempts total 6.63 hours, including failures and superseded completed attempts, with a largest sampled process-tree RSS of 1,845 MiB. These records exclude interrupted or standalone diagnostic work and are not total project time. All primary SARIMAX fits complete in this implementation. Memory requirements depend on implementation and environment, as well as model specification.

\section{\texorpdfstring{Discussion and future work}{Discussion and future work}}\label{sec:discussion}

The comparison shows that updating all model parameters each week does not provide the same benefit for every family. SARIMAX's frozen and refitted aggregate errors are close, whereas Prophet and the seed-42/123 NeuralProphet runs improve with refitting. An established scalar EWMA correction also reduces their persistent error component, without a repeated parameter fit. Its smaller SARIMAX benefit is uncertain in the paired weekly analysis, and its adverse effect on NeuralProphet seed 2026 shows that adaptation can overcorrect an already competitive base forecast.

These findings support a conditional comparison of update policies, rather than a universal preference for statistical or neural architectures. NeuralProphet's performance varies across seeds and improves relative to the other families on the small high-concentration subset. Broader-input controls also show that predictor availability can change errors more than the choice of update strategy. Model interpretability is limited to fitted associations and decomposed forecasts; correlated gases and API-derived targets prevent causal conclusions.

Computationally, correction arithmetic is inexpensive and avoids repeated parameter optimization. However, frozen SARIMAX still refreshes its state, NeuralProphet still consumes recent history, and all models require future gas inputs under the present protocol. Differing fit timers and sampled memory measurements prevent a single controlled cross-model speed claim. Deployment would require externally forecast or otherwise available covariates and a prospective evaluation of their errors.

Future work should first examine forecast-available gas inputs and independently verified station targets over additional seasons and locations. Longer chronological validation and additional seeds could assess the stability of settings and correction behavior. Attention, temporal/cross-pollutant transfer, and distributed learning offer complementary extensions \citep{revSTA,revWinter,revCrossPollutant,bib30}; they require evaluation under matched information sets before claims of improved accuracy or efficiency.

\section{Limitations}\label{sec:limitations}

There are a number of caveats to be noted. First, the data are an API estimate series, not station ground truth. Original response archives and retrieval dates are unavailable, pollutant field mapping cannot be independently verified, and the source-start discrepancy remains unresolved. Predicting the given series does not demonstrate accuracy against physical measurements.

Second, missing timestamps, four invalid predictor values, and partial-week coverage must be handled explicitly. Causal seasonal filling provides inference history only, but its assumptions may be weak over long gaps, including a 120-hour interval. The observed-hour mask does not establish accuracy during missing periods. Primary targets are not clipped, and high-event findings use only 158 hours concentrated in six weeks.

Third, actual future covariates make this a retrospective Perfect Prognosis experiment. The information set is different for target-only baselines, and future PM$_{10}$ benefits the broader controls. The broad-input SARIMAX control is unavailable after nonconvergence. Four weeks of winter validation, three NeuralProphet seeds, and 23 test origins from January to June limit seasonal and geographic generalization. Earlier examination of the test period also limits confirmatory claims.

Finally, the bounded grid does not establish optimal settings, solver convergence does not guarantee global optimality, and short-series bootstrap intervals are descriptive. Conditional coefficients and components do not represent causal effects. Runtime boundaries vary and memory is sampled, so no universal computational rankings or production-level speedups can be drawn.

\section{Conclusion}\label{sec:conclusion}

This study investigated hourly PM\textsubscript{2.5} prediction in Beijing using SARIMAX, Facebook Prophet, and NeuralProphet with weekly refitting and frozen parameters with EWMA correction. The workflow preserves original observed targets, uses training-only validation and transformations, and compares all primary streams on the same 23 origins and 3,624 observed hours under Perfect Prognosis.

SARIMAX gives the lowest aggregate errors in the selected comparison using four gases, while refitting and residual correction offer model- and seed-dependent benefits. NeuralProphet's seed variability and performance at high concentrations, together with improvements from a wider set of predictors, limit a universal family ordering. Conditional interpretations are given by explicit regressor effects and component summaries, and isolated correction timings indicate low arithmetic overhead. These results describe adaptation in the evaluated information setting. Further testing with gas inputs available at the time of the forecast and authenticated station measurements is needed before operational conclusions.

\section*{Data availability}\label{sec:data_availability}

The dataset used in this study is included in the publicly available Zenodo archive: \url{https://doi.org/10.5281/zenodo.23170178}.

\backmatter

\bibliography{references}

\section*{Statements and Declarations}

\bmhead{Funding}

The authors declare that no funds, grants, or other support were received during the preparation of this manuscript.

\bmhead{Competing interests}

The authors have no relevant financial or non-financial interests to disclose.

\bmhead{Author contributions}

All authors contributed to developing the study methodology and experimental direction through regular discussions. Moazzam Umer Gondal prepared the data, developed and executed the code, conducted the experiments, analyzed and visualized the results, and wrote the original manuscript. Hamad Ul Qudous and Asma Ahmad Farhan supervised the research, provided substantial methodological guidance, helped refine the experimental design and subsequent research steps, and contributed to the interpretation of the results. Hamad Ul Qudous and Asma Ahmad Farhan critically reviewed the manuscript. All authors discussed the findings and read and approved the final manuscript.

\bmhead{Ethics approval}

Not applicable.

\bmhead{Consent to participate}

Not applicable.

\bmhead{Consent for publication}

Not applicable.

\bmhead{Materials availability}

Not applicable.

\end{document}